# Adversarial construction as a potential solution to the experiment design problem in large task spaces

Prakhar Godara (prakhargodara@gmail.com)[1], Frederick Callaway[1] & Marcelo G. Mattar[1,2]
[1]Department of Psychology, New York University
[2]Center for Neural Science, New York University

**Abstract**

Despite decades of work, we still lack a robust, task-general theory of human behavior even in the simplest domains. In this paper we tackle the generality problem head-on, by aiming to develop a unified model for all tasks embedded in a task-space. In particular we consider the space of binary sequence prediction tasks where the observations are generated by the space parameterized by hidden Markov models (HMM). As the space of tasks is large, experimental exploration of the entire space is infeasible. To solve this problem we propose the adversarial construction approach, which helps identify tasks that are most likely to elicit a qualitatively novel behavior. Our results suggest that adversarial construction significantly outperforms random sampling of environments and therefore could be used as a proxy for optimal experimental design in high-dimensional task spaces.

## Introduction

A central goal of cognitive science is to characterize the algorithms that underlie human learning and decision-making. Decades of research have produced detailed models of behavior in classic paradigms, from probability matching in binary prediction tasks to exploration–exploitation tradeoffs in multi-armed bandits. Yet these successes have not cumulated into a unified theory. Models that fit one paradigm often fail in another, and even within a paradigm, fitted parameters can shift substantially as superficial task features change (Eckstein et al., 2022). This fragility suggests that single-task modeling, however refined, may be insufficient to recover the underlying cognitive algorithm (Almaatouq et al., 2024).

If behavior on any single task reveals only a narrow slice of the cognitive algorithm, the natural remedy is to study behavior across a *task space*—a parametric family of environments that spans qualitatively different demands. In principle, fitting a single model to data from many tasks should yield a more complete and robust characterization. In practice, however, this approach faces a severe methodological barrier: since tasks can vary along many dimensions, even modest task families grow too large to explore exhaustively. Random sampling of tasks, though unbiased, is susceptible to wasting experimental budget on environments that are either redundant with each other or weakly diagnostic of the behavioral assumptions under test (Wilson & Collins, 2019). The result is that behavioral modeling across a task space remains more aspiration than methodology.

The challenge for a cognitive scientist is, then, one of experiment design: which tasks should we select to learn the most about the underlying behavioral algorithm? One might consider Bayes-optimal experiment design; however, for continuous task spaces larger than a few dimensions, determining the most informative experiment becomes computationally intractable, setting aside the further difficulty of specifying prior distributions over participant behavior (Ryan et al., 2016). A practical way forward therefore requires a principled *heuristic*.

A useful clue comes from the rational analysis tradition, which holds that human behavior is often well-adapted to task demands (Anderson, 1990; Griffiths et al., 2015). If this is true, and human behavior remains well-adapted even in tasks where a given model performs poorly, then the most informative experiments are those in which our current model diverges most from what a well-adapted agent would do—that is, the tasks that best expose the model's inadequacy. The same conclusion follows from a complementary perspective: if participants possess a repertoire of strategies and deploy whichever one suits the current task, then observing behavior in environments where a previously successful strategy fails necessarily forces the deployment of a different strategy. Therefore, if we iteratively select tasks that maximize regret, we should efficiently force novel behaviors to emerge and thereby accelerate the discovery of a task-general account. Despite the appeal of this logic, no existing method implements it for cognitive task spaces, and it remains unknown whether regret-driven task selection actually outperforms simpler alternatives in human experiments.

In this paper, we propose *adversarial construction* (AC): an iterative procedure that fits a behavioral model to data collected so far and then selects the next task to maximally expose the model's limitations. The rationale behind AC is to identify tasks that incentivize humans to behave in novel ways relative to previous tasks, thereby allowing a broader characterization of the underlying behavioral algorithm. We evaluate AC in a binary sequence prediction task where environments are parameterized by two-state hidden Markov models—a simple yet fundamental domain that admits a well-defined normative baseline. In human experiments, we find that AC converges within approximately six iterations to a small set of behaviorally critical task types, including change detection, rare-event detection, and cyclic alternation. Models trained on AC-selected data achieve substantially lower worst-case generalization error than models trained on randomly sampled tasks and better predict behavior in novel tasks selected to have structure not reflected in the training set—despite using less data. Finally, interpretability analyses reveal that the AC-trained model acquires internal representations aligned with bigram statistics, the minimal state representation required for this task family. Together, these results suggest that adversarial construction offers a viable path toward efficient, task-general characterizations of human cognition.

## Adversarial construction

Consider an agent interacting with an environment over a horizon $T$. At each time $t \in \{1, \ldots, T\}$, the agent chooses an action $a_t \in \mathcal{A}$ and the environment returns an outcome $o_t \in \mathcal{O}$.[1] Let the history up to time $t$ be $h_t := (a_1, o_1, \ldots, a_t, o_t)$. An agent is a (possibly stochastic) policy $\pi : \mathcal{H} \to \Delta(\mathcal{A})$, and an environment is a (possibly nonstationary) kernel $\mathcal{M} : \mathcal{H} \times \mathcal{A} \to \Delta(\mathcal{O})$.

Let $\mathfrak{T}$ be a *task space*, i.e., a nonempty set of admissible environments $\mathfrak{T} \subseteq \{\mathcal{M} : \mathcal{H} \times \mathcal{A} \to \Delta(\mathcal{O})\}$, and $J(\pi, \mathcal{M}) \in \mathbb{R}$ be the performance functional. Given a prior over tasks $\rho \in \Delta(\mathfrak{T})$, the optimal policy is defined

$$J_\rho(\pi) := \mathbb{E}_{\mathcal{M} \sim \rho}[J(\pi, \mathcal{M})], \qquad \pi_\rho^* = \arg\max_{\pi \in \Pi} J_\rho(\pi),$$

and the (Bayes-relative) regret on a single environment $\mathcal{M}$ as

$$d_\rho(\pi, \mathcal{M}) := J(\pi_\rho^*, \mathcal{M}) - J(\pi, \mathcal{M}).$$

Given a candidate behavioral model $\pi$, *adversarial construction* selects a task on which $\pi$ performs worst relative to the reference ($\pi^*$):

$$\mathcal{M}_\pi \in \arg\max_{\mathcal{M} \in \mathfrak{T}} d(\pi, \mathcal{M}). \tag{1}$$

Iterating the procedure—collecting data, refitting the model on the aggregate data, and selecting the next adversarial task describes the adversarial construction loop (c.f. Figure 2 for a schematic representation).

## Setup and implementation

### Task

We consider a binary sequence prediction task, where the action space $\mathcal{A} = \{0, 1\}$ represents the agent's possible predictions at a given time $t$ and $\mathcal{O} = \{0, 1\}$ represents its observations at time $t$. We consider a finite horizon task with $T = 50$. The agent's goal is to maximize the number of correct predictions, i.e., $\sum_t^T \mathbf{1}\{a_t = o_t\}$.

We restrict the task space $\mathfrak{T}$ to a parametric family of two-state hidden Markov models (HMMs). Let $s_t \in \{1, 2\}$ denote the latent state, with initial distribution $\mu \in \Delta(\{1, 2\})$. The states transition at each time step according to the transition matrix $P(s_{t+1}|s_t)$, which is given by

$$P(s_{t+1}|s_t) = \begin{cases} p_i, & s_t = i \neq s_{t+1}, \\ 1 - p_i, & s_t = i = s_{t+1}. \end{cases} \tag{2}$$

[1] The outcome $o_t$ may include rewards, observations, feedback, or other task-specific signals.

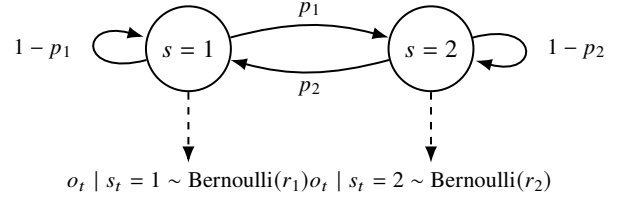

Figure 1: Two-state hidden Markov model. From state 1 the chain switches to state 2 with probability $p_1$ (otherwise stays with $1 - p_1$); from state 2 it switches to state 1 with probability $p_2$ (otherwise stays with $1 - p_2$). Outcomes are emitted as Bernoulli with parameters $r_1$ and $r_2$ respectively.

Given a state, the HMM emits a binary observation $o_t \in \mathcal{O}$ with emission probabilities $r_i$, i.e.,

$$P(o_t = 1 | s_t = i) = r_i. \tag{3}$$

A task is therefore fully specified by the parameters $(\mathbf{r}, \mathbf{p}, \mu)$[2] and is visually depicted in Figure 1 where $\mathbf{r} = (r_1, r_2)$ and $\mathbf{p} = (p_2, p_2)$.

### Policy class and objective

We fit recurrent policies $\pi_\theta : \mathcal{H} \to \Delta(\mathcal{A})$ implemented by a single-layer Gated Recurrent Unit (GRU, Cho et al., 2014) with hidden dimension $d = 7$. Using the feature vector $x_t = (a_{t-1}, o_{t-1})$, the network updates

$$z_t = \text{GRU}_\theta(z_{t-1}, x_t), \qquad \ell_t = w^\top z_t + b,$$

and predicts via

$$\pi_\theta(a_t = 1 \mid h_{t-1}) = \sigma(\ell_t), \qquad \sigma(u) = \frac{1}{1 + e^{-u}}.$$

We evaluate a policy $\pi$ on $\mathcal{M}$ by expected prediction accuracy,

$$J(\pi, \mathcal{M}) = \mathbb{E}_{\pi, \mathcal{M}} \left[ \frac{1}{T} \sum_{t=1}^{T} \mathbf{1}\{a_t = o_t\} \right].$$

### Ideal learner baseline and regret

As a normative reference, we use an ideal learner $\pi^{\text{IL}}$ that performs approximate Bayesian filtering in the same two-state HMM family using a Rao-Blackwellized particle filter (Doucet et al., 2001). The ideal learner knows the number of states and the initial distribution $\mu$, but not the transition or emission parameters, which it infers online (with a prior favoring deterministic parameters). At time $t$, it forms $p_t^{\text{IL}} = \Pr(o_t = 1 \mid h_{t-1})$ and samples $a_t \sim \text{Bernoulli}(p_t^{\text{IL}})$. Using this baseline, we define the regret as

$$d(\pi, \mathcal{M}) := J(\pi^{\text{IL}}, \mathcal{M}) - J(\pi, \mathcal{M}), \tag{4}$$

which we use to perform AC.

[2] Throughout this work we fix $\mu = (0.5, 0.5)$.

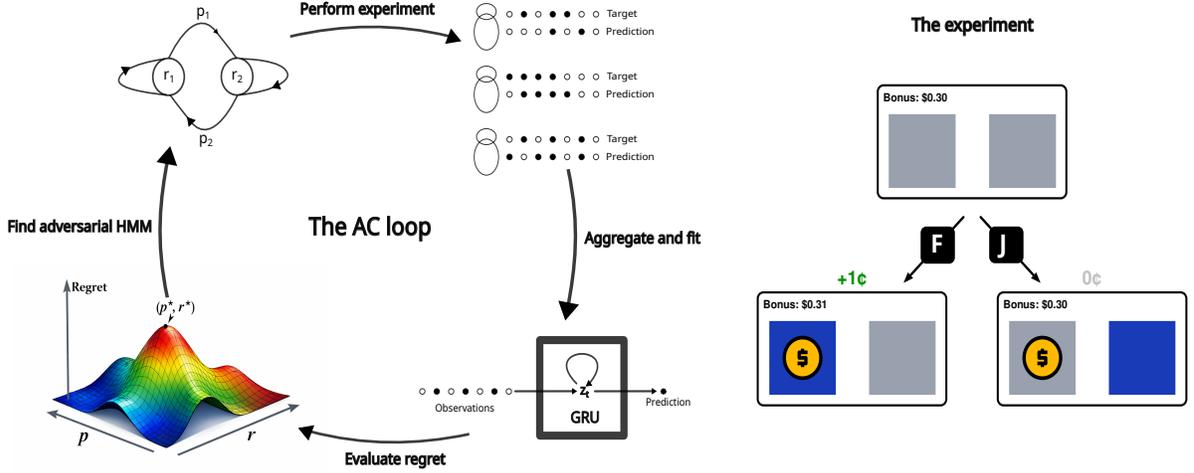

Figure 2: Schematic of adversarial construction loop (left): select an environment → conduct the experiment → fit a policy to data → compute the regret → maximize the regret to obtain the adversarial environment and (right) the experiment task presented as a coin game.

**Human experiment**

A total of 1400 participants performed the binary sequence-prediction task on a small set[3] of environments $\mathcal{M} \in \mathfrak{T}$, selected either by adversarial construction (AC) or by uniform random sampling from $\mathfrak{T}$. Each participant performed only in a single environment. Participants received only minimal task instructions (how to enter predictions and how accuracy affected reward); no information about latent states or HMM structure was provided. We applied no exclusion criteria to preserve naturalistic human behavior.

100 participants completed a task with i.i.d. observations, obtained by setting $r_1 = r_2 = r$ with $r \sim \text{Unif}[0, 1]$. This shared dataset, denoted $\mathcal{D}_1$, serves as the initialization for AC and as the first dataset for all corpora.

**AC dataset.** AC then selected five additional environments $\mathcal{M}_2^{AC}, \ldots, \mathcal{M}_6^{AC}$, producing behavioral datasets $\mathcal{D}_2^{AC}, \ldots, \mathcal{D}_6^{AC}$. The AC corpus is then given by $\mathcal{D}_{1:6}^{AC} := \mathcal{D}_1 \cup \bigcup_{g=2}^{6} \mathcal{D}_g^{AC}$.

**Random-sampling datasets.** Independently, we sampled 8 environments $\{\mathcal{M}_j^R\}_{j=1}^8$ uniformly by drawing $(r_1, r_2, p_1, p_2) \sim \text{Unif}[0, 1]^4$, and collected the corresponding datasets $\mathcal{D}_1^R, \ldots, \mathcal{D}_8^R$. We formed $K = 10$ random subsets $S_k \subset \{1, \ldots, 8\}$ with $|S_k| = 5$. For each $k$, we constructed a random corpus by combining $\mathcal{D}_1$ with the five selected random datasets, i.e.

$$\mathcal{D}_{1:6}^{R,k} := \mathcal{D}_1 \cup \bigcup_{j \in S_k} \mathcal{D}_j^R, \qquad k = 1, \ldots, 10.$$

---
[3]A total of 14 environments were chosen with 100 participants per environment.

**Model fitting.** For each corpus $\mathcal{D}_{1:6}^*$ (where $*$ indexes either AC or a random corpus), we trained 6 GRU policies by fitting one model per cumulative prefix: for $i = 1, \ldots, 6$, we fit $\pi_i^*$ on $\mathcal{D}_{1:i}^* := \bigcup_{g=1}^{i} \mathcal{D}_g^*$, giving us a total of 66 models.

## Results

We evaluated adversarial construction (AC) against random task sampling in a binary sequence prediction task where environments were parameterized by a two-state hidden Markov model. As aforementioned, both approaches began with the same seed dataset $\mathcal{D}_1$ collected on the i.i.d. task, then iteratively expanded their corpora: AC selected each subsequent task to maximize predicted regret, while the random baseline sampled tasks uniformly from the parameter space. At each iteration, we fit a GRU policy to the cumulative dataset and evaluated generalization across the full set of collected environments. Our analyses address four questions: (1) Does AC converge to a stable set of diagnostic tasks? (2) Do models trained on AC-selected data generalize better than those trained on randomly sampled data? (3) What types of environments does AC select, and why are they diagnostic? (4) Does the AC-trained model acquire appropriate internal representations for this task family?

### AC identifies a small set of behaviorally critical tasks

AC rapidly converges to a small set of diagnostic task regions. To track this convergence, we distinguish three quantities: the *predicted* regret (the regret of the current model $\pi_n$ on the selected task, computed before data collection), the *observed* regret (the gap between human accuracy and ideal-learner accuracy on that task), and the *postdicted* regret (the regret of the refit model $\pi_{n+1}$ on the same task). Across all iterations, AC reliably targeted tasks with large observed regret

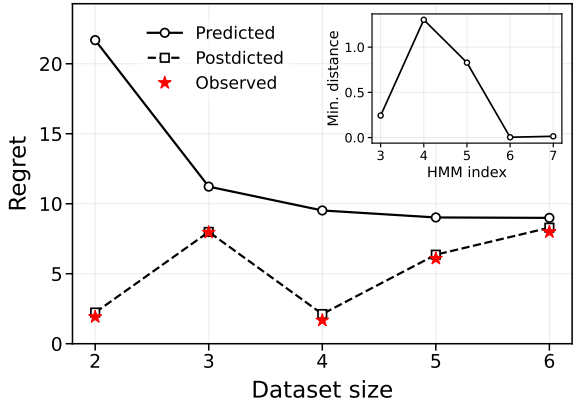
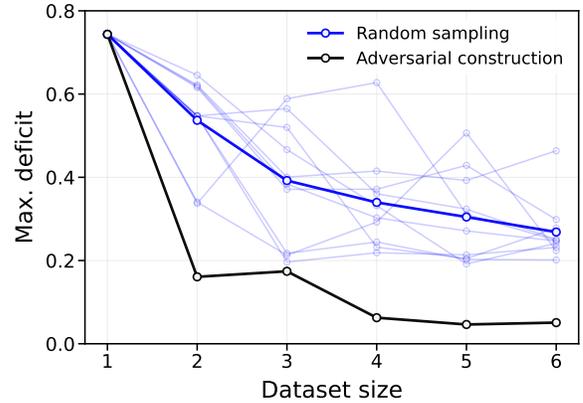

Figure 3: AC converges within approximately six iterations. *Main panel:* Predicted, postdicted, and observed regret on adversarially selected tasks as a function of iteration. Predicted regret consistently exceeds observed regret, confirming that humans outperform model predictions on adversarial tasks. *Inset:* Minimum distance between each newly selected HMM and previously selected HMMs; rapid stabilization indicates convergence to a small set of diagnostic task types.

Figure 4: AC yields faster generalization than random sampling. Worst-case fit deficit (maximum NLL gap relative to best model, computed over held-out environments) as a function of the number of training environments for the AC corpus (black) and the ten random corpora (light blue, with mean dark blue). AC achieves lower worst-case deficit than random sampling at every dataset size.

(Figure 3). Notably, predicted regret consistently exceeded observed regret—that is, humans always outperformed the current model's predictions on adversarial tasks. This pattern validates the core assumption behind AC: adversarial tasks expose model limitations precisely because humans possess strategies the model has not yet captured. As iterations progressed, predicted and postdicted regrets converged, indicating that the AC loop progressively reduced systematic error. By the sixth iteration, this gap nearly vanished, signaling convergence.

A complementary measure of convergence confirms this pattern. We quantified how "novel" each newly selected environment was relative to previously selected ones by computing the minimum symmetry-reduced distance between each new HMM and the set of prior HMMs.[4] This distance decreased rapidly and stabilized within a few iterations (Figure 3, inset), confirming that AC quickly concentrates on a small set of behaviorally critical regions within the task space $\mathfrak{T}$.

### AC accelerates task-generalization relative to random sampling

We now move to our key result: How well do the learned models generalize across tasks in $\mathfrak{T}$? To quantify the generalization capabilities of our models, we evaluate each fitted model by its mean negative log-likelihood (NLL) on behavioral data on the entire corpus of 14 environments we collected, i.e. $\mathcal{D}_{1:6}^{R,k} := \mathcal{D}_1 \cup \bigcup_{i \in \{2,\cdots,6\}} \mathcal{D}_i^{AC} \cup \bigcup_{i \in \{1,\cdots,8\}} \mathcal{D}_i^{R}$. We then define for each model the *fit deficit* on a given dataset $\mathcal{D}_i^*$ which is given by the difference of its mean NLL score on the dataset from the mean NLL score of the best performing model[5] on the same dataset.

We found that models trained on AC-selected data achieved substantially lower worst-case deficit than those trained on randomly sampled data, and did so with fewer training environments (Figure 4). This advantage emerged early and persisted throughout training, indicating that AC produces task-general behavioral models more efficiently than random sampling.

### AC-trained models capture trial-by-trial behavioral dynamics

Aggregate generalization metrics do not guarantee that models capture the fine-grained dynamics of human behavior. To assess trial-by-trial fit, we compared model predictions to human choice trajectories on two specific observation sequences generated from environments outside the training corpus.[6] For each sequence, we computed the average human choice probability at each time step and compared it to the predictions of two models: the GRU trained on the AC corpus and the GRU trained on the full randomly sampled corpus (which contained 33% more data than the AC corpus).

Despite the data disadvantage, the AC-trained model tracked human choice dynamics more closely than the random-trained model on both sequences, as reflected in lower $L_1$ distances between model and human trajectories (Figure 5). These results confirm that the generalization advantage of AC extends beyond aggregate metrics to the trial-level behavioral signatures that are critical for cognitive modeling.

---

[4] Distance is computed in the four-parameter representation $(p_1, p_2, r_1, r_2)$ after minimizing over the natural symmetries of the two-state HMM (state relabeling and emission relabeling).

[5] Note that this model could also be one of the models that is trained on the behavior observed in this environment. Therefore the deficit is being measured relative to an approximate performance ceiling on the dataset.

[6] The environments and sequences were semi-manually selected to have structure not present in any AC dataset: (1) switch-detection with occasional false alarms and (2) alternation with occasional repetitions. See black and white circles in Figure 5.

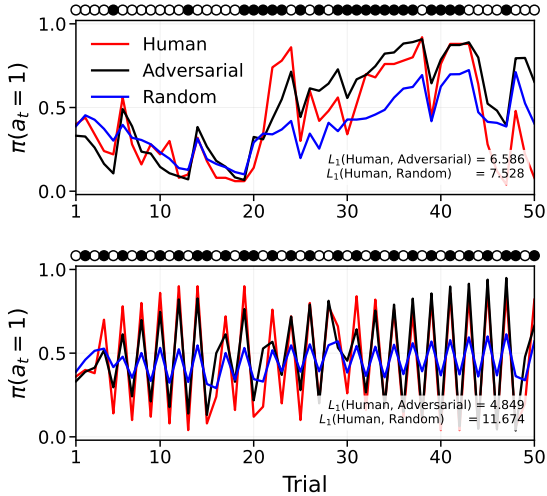
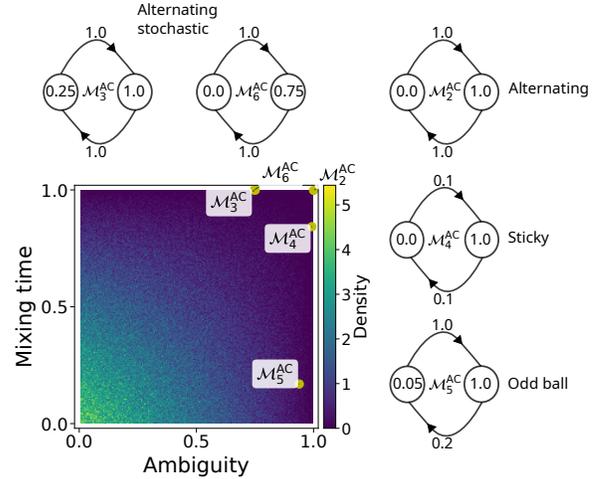

Figure 5: AC-trained models better capture trial-by-trial human behavior. Average choice probability ($\pi(a_t = 1)$) for humans (black), the AC-trained GRU (red), and the GRU trained on the full random corpus (blue) on two held-out observation sequences (shown at top; solid circles = 1, hollow circles = 0). $L_1$ distances between model and human trajectories are reported for each comparison. The AC-trained model achieves lower $L_1$ distance on both sequences despite being trained on 33% less data.

Figure 6: AC selects qualitatively distinct, diagnostic environments. Background shading shows the density of HMM types under uniform random sampling, plotted by mixing time and ambiguity. Colored points (yellow) indicate the specific environments selected by AC, along with their HMM representations. Random sampling concentrates on high-ambiguity, fast-mixing environments; AC targets low-ambiguity environments with non-trivial dynamics.

**AC selects diagnostic, learnable environments**

Why does AC yield better generalization than random sampling? One possibility is that AC selects qualitatively distinct environment types that collectively span the behaviorally relevant regions of the task space. To test this, we compared the distribution of environments selected by AC to those obtained by uniform random sampling (Figure 6). Random sampling frequently yielded environments with small *mixing time* (rapid approach to stationarity) and high *ambiguity* (similar emission rates across states)—environments that are often redundant or weakly diagnostic. In contrast, AC concentrated on environments with non-trivial transient dynamics that remained predictable for the ideal learner yet were maximally challenging for the current fitted model. These tasks are simultaneously learnable and diagnostic, explaining AC's data efficiency.

The specific sequence of AC-selected environments illustrates this point. The first adversarial environment, $\mathcal{M}_2^{AC}$, was a near-deterministic alternating sequence of 0s and 1s—a task that defeats the i.i.d. strategy from the seed task, which relies only on bit frequency. The second, $\mathcal{M}_3^{AC}$, also alternated deterministically between states but introduced stochastic emissions in one state; success here requires tracking 2-bit (bigram) frequencies rather than a simple win-shift rule. The third, $\mathcal{M}_4^{AC}$, featured stable regimes with rare switches, introducing a more gradual decay of autocorrelation. The fourth, $\mathcal{M}_5^{AC}$, favored one bit most of the time but included rare, independent flips—an "oddball" structure. Finally, $\mathcal{M}_6^{AC}$ was an emission-reversed version of $\mathcal{M}_3^{AC}$, at which point AC had effectively converged. Each environment in this sequence was adversarial to the strategies that succeeded on previous environments, forcing the model to capture increasingly general aspects of human behavior.

**AC yields models with appropriate internal representations**

Does AC drive the learnt model toward internal representations that are appropriate for the task space? To answer this question, we compared the GRU trained only on the i.i.d. seed task ("initial") to the GRU obtained after AC convergence ("final") using two complementary analyses.

First, we performed a distillation analysis to identify the minimal state representation captured by each GRU. We fit each GRU's policy using generalized linear models (GLMs) that included *n*-gram Q-values—exponentially weighted recency scores for the occurrence of *n*-grams—along with run length, recent outcomes, and their interactions (see Figure 7 caption for details). We fit three GLM variants corresponding to unigram ($n = 1$), bigram ($n = 2$), and trigram ($n = 3$) state representations. For the initial GRU, the unigram model provided a reasonable fit. As AC progressed, however, the unigram fit quality degraded substantially, while the bigram model maintained high fit quality throughout (Figure 7, top). The trigram model offered little improvement over the bigram despite having more parameters. This pattern indicates that AC drives the model toward a bigram representation—the minimal state representation sufficient for optimal performance in this HMM family.

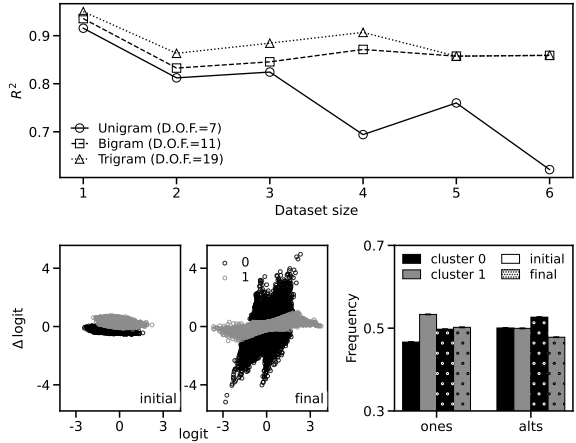

Figure 7: AC drives the model toward bigram representations. *Top:* Fit quality ($R^2$) of GLM distillations using unigram, bigram, and trigram Q-learning features as a function of AC iteration. The unigram model's fit degrades over iterations; the bigram model remains stable; the trigram model provides little additional benefit. *Bottom:* Clustering of length-15 sequences by GRU logits for the initial model (trained on i.i.d. data only) and the final model (after AC convergence). The initial GRU clusters sequences by bit frequency ("ones"); the final GRU clusters by alternation frequency ("alts").

Second, we examined the GRU's internal representations directly by clustering observation sequences according to the GRU's hidden-state dynamics. We generated all possible sequences of length 15 and clustered them based on the GRU's logits at the final two time steps using a Gaussian mixture model. The initial and final GRUs produced qualitatively different cluster structures (Figure 7, bottom). For the initial GRU, the two clusters differed primarily in mean bit frequency (the proportion of 1s). For the final GRU, the clusters were instead differentiated by alternation frequency (the proportion of bit switches), with similar mean bit frequencies across clusters. This shift from unigram to bigram features confirms that AC drives the model toward the state representation demanded by the task space.

## Discussion

Our results establish that adversarial construction (AC) is a viable approach for efficiently selecting tasks in a continuous task space for task-general behavioral modeling. In a binary sequence prediction task parameterized by two-state HMMs, we found that AC converged within approximately six iterations to a small set of diagnostic environment types, including deterministic alternation, stochastic alternation, stable regimes with rare switches, and oddball structures. Models trained on AC-selected data achieved substantially better worst-case generalization than models trained on randomly sampled data, and did so with less data. Interpretability analyses confirmed that AC drove the model toward bigram representations, the minimal state representation sufficient for this task family. Together, these results establish AC as a viable heuristic for task-space experiment design.

The present work focused on a minimal yet fundamental task family—passive two-state HMMs—and a central next step is to scale AC to richer domains. One axis of extension is increasing latent complexity: more than two states, hierarchical or factorial latent structure, and nonstationary dynamics. Another axis is moving from passive observation to *active* environments where actions influence state transitions (i.e., stochastic finite state automata (SFSA) or POMDP-like settings). These extensions introduce qualitatively new phenomena, such as exploration–exploitation tradeoffs, strategic information gathering, and planning over beliefs, and would provide a sharper test of whether AC can identify task regimes that force distinct human strategies to emerge.

A second limitation concerns data efficiency in high-dimensional task spaces. Adversarially selected environments in the present setting tended to be near-deterministic, which is desirable because it creates steep performance landscapes and amplifies differences between behavioral hypotheses. However, highly predictable environments generate low-entropy trajectories, making successive trials—and often successive participants—partially redundant. As task dimensionality grows, this redundancy could become a bottleneck. A promising solution is to incorporate an explicit *diversity* or *informativeness* regularizer into the selection objective, for example penalizing environments with low outcome entropy, low Fisher information, or high similarity to previously sampled tasks. More generally, AC can be viewed as trading off two desiderata: selecting environments that maximize diagnostic failure of the current model, while ensuring sufficient variability in the observed data to support parameter-efficient learning.

In sum, adversarial construction offers a practical route toward studying human cognition at the level of task spaces rather than individual tasks. Realizing its full potential will require scaling to richer and more interactive task families, improving data efficiency through diversity-aware selection, and developing theoretical foundations that clarify when and why regret-driven experiment design succeeds. These advances would bring cognitive science closer to the goal of characterizing the algorithms underlying human learning and decision-making in their full generality.